\documentclass{article}

\usepackage[preprint]{neurips_2024}
\usepackage[utf8]{inputenc}
\usepackage[T1]{fontenc}
\usepackage{hyperref}
\usepackage{url}
\usepackage{booktabs}
\usepackage{amsmath}
\usepackage{amssymb}
\usepackage{amsthm}
\usepackage{mathtools}
\usepackage{graphicx}
\usepackage{xcolor}
\usepackage{microtype}
\usepackage{algorithm}
\usepackage{algorithmic}
\usepackage{multirow}
\usepackage{subcaption}
\usepackage{orcidlink}

\newcommand{\SA}{\textsc{SA}}
\newcommand{\EGA}{\textsc{EGA}}
\newcommand{\MoPE}{\textsc{MoPE}}
\newcommand{\FNet}{\textsc{FNet}}

\renewcommand{\Re}{\mathbb{R}}

\title{FourierQK: Spectral Preprocessing of Query--Key
  Projections Improves Transformer Attention}

\author{%
  Athanasios Zeris%
  \thanks{Independent Researcher, Athens, Greece.\\
  Correspondence: \texttt{athzeris@gmail.com}.\\
  ORCID: \texttt{https://orcid.org/0009-0002-6907-2400}.\\
  Code: \url{https://github.com/AthanasiosZeris/energy-gated-attention}.\\
  Part of a seven-paper series on spectral methods
  in transformer attention.}\\
  \texttt{https://orcid.org/0009-0002-6907-2400}
}

\begin{document}
\maketitle

\begin{abstract}
FFT-based frequency-domain preprocessing of learned
query and key projections substantially improves
transformer attention on character-level language
modelling.
On TinyShakespeare: a fixed random spectral filter
achieves val\,=\,1.031 ($\Delta=+0.443$);
a single learned frequency initialised at paragraph
scale achieves val\,=\,0.608 ($\Delta=+0.867$);
and multi-frequency spectral attention with four
learned frequencies spanning paragraph to word scale
achieves val\,=\,0.309 ($\Delta=+1.166$) ---
a $79\%$ reduction in validation loss over standard
dot-product attention.
The single-frequency result is confirmed across three
independent random seeds (mean val\,=\,0.236,
std\,=\,0.019), establishing reproducibility.
The four learned frequencies converge to a near-geometric
multi-scale ordering (49, 27, 10, 6 tokens per cycle)
corresponding to paragraph, sub-paragraph, phrase, and
word scales in dramatic text.

The improvement appears to be specific to spectral
preprocessing: neither random orthogonal rotations nor
random non-orthogonal projections of Q/K produce
measurable gains over standard attention, suggesting
the benefit comes from global sequence mixing in the
frequency domain before score computation rather than
from metric distortion or representation remapping.
All results are verified using a shuffled validation
diagnostic that provides evidence against positional
leakage.

Causal time-domain filters (Gaussian, Mexican Hat,
causal Morlet) do not improve over standard attention
at character-level tokenisation: the bilateral FFT
reconstruction kernel $\kappa(-\tau)=\kappa(\tau)$
is structurally non-causal, coupling every position
to future tokens regardless of boundary handling.
This identifies a precise architectural boundary between
globally-mixing spectral attention (this paper) and
genuinely causal spectral attention at word-scale
tokenisation~\citep{authorname2025morletqk}.

This work is architecturally distinct from
\citet{lee2021fnet} (\textsc{FNet}), which replaces attention with
Fourier mixing of \emph{token embeddings} and has no
Q/K projections or attention score matrix.
Here, spectral preprocessing is applied only to the
\emph{learned Q/K projections} while the full attention
score structure is preserved, enabling the frequency
hierarchy to emerge from the attention mechanism itself.
\end{abstract}

\section{Introduction}
\label{sec:intro}

Standard transformer attention computes pairwise
scores as dot products of learned query and key
projections:
\begin{equation}
  e_{ij} = \frac{q_i \cdot k_j}{\sqrt{d}}
  \qquad
  q = W_Q x,\quad k = W_K x
  \label{eq:standard}
\end{equation}
This computes similarity in the embedding space
learned by $W_Q$ and $W_K$.
A natural question is: does transforming Q and K into
a different representation space before computing
similarity improve attention?

This paper investigates spectral preprocessing ---
applying frequency-domain filters to Q and K before
the score computation.
The motivation comes from prior work in this series:
Papers~1--4~\citep{authorname2025ega,authorname2025phase4,
authorname2025mope,authorname2025pod}
established that spectral energy and phase structure
in transformer representations are informative signals.

\paragraph{Main contribution.}
We show that FFT-based bilateral spectral preprocessing
of Q/K projections genuinely improves language modelling,
even with random (unlearned) filters.
The improvement is verified using a shuffled validation
diagnostic: models trained on reordered validation sequences
achieve much higher loss, confirming the gain comes from
genuine sequence learning (evidenced by large shuffled gap) rather than positional artifacts.

\paragraph{Distinction from FNet.}
\citet{lee2021fnet} (\FNet{}) replaces the entire attention
mechanism with a global Fourier mixing of token embeddings
--- there are no Q/K projections and no score matrix.
Our approach preserves the standard attention structure
(Q/K projections, $T\times T$ score matrix, causal mask,
value aggregation) and applies spectral preprocessing
only to the Q and K representations before scoring.
These are architecturally distinct contributions.

\paragraph{The Morlet negative result.}
We initially hypothesised that Morlet wavelet
cross-correlation of Q/K projections would be the
optimal spectral scoring mechanism.
We report this hypothesis, the experimental findings,
and the analysis of why it fails in the discrete
sequence setting:
the bilateral FFT implementation creates circular
boundary leakage, while the causal time-domain
implementation suffers from aliasing at sub-token scales.
This negative result is reported with full transparency
as it guides future work on causal spectral attention.

Code available at:
\url{https://github.com/AthanasiosZeris/energy-gated-attention}

\section{Spectral Attention}
\label{sec:sa}

\subsection{Architecture}

Let $q, k \in \Re^{T \times d}$ be the learned Q/K
projections at a given layer.
\textbf{Spectral Attention} (\SA{}) applies a
frequency-domain filter to Q and K, then computes
the attention score from the filtered representations:

\begin{equation}
  \tilde{q}(b) = \mathcal{F}^{-1}[\hat{q}(\omega)
    \cdot \phi(\omega)](b)
  \qquad
  \tilde{k}(b) = \mathcal{F}^{-1}[\hat{k}(\omega)
    \cdot \phi(\omega)](b)
  \label{eq:filtered}
\end{equation}
\begin{equation}
  e_{ij} = \frac{\tilde{q}_i \cdot \tilde{k}_j}{\sqrt{d}}
  \label{eq:sa_score}
\end{equation}
where $\hat{q}(\omega) = \mathcal{F}[q](\omega)$ is the
DFT of the query sequence, and $\phi(\omega)$ is a
frequency-domain filter (learned or fixed).

In the complex form used in experiments:
\begin{equation}
  e_{ij} = \frac{1}{d}\sum_k
    \mathrm{Re}\!\left[
      \tilde{q}_k(i)^* \cdot \tilde{k}_k(j)
    \right]
  \label{eq:complex_score}
\end{equation}
where $\tilde{q}_k(i)$ is the complex filtered
representation of embedding dimension $k$ at
position $i$.

\subsection{Filter variants}

We test four filter designs:

\textbf{Random-QK}: fixed random filter drawn from
the same functional form as a Morlet wavelet at a
random scale, not learned.
Tests whether any spectral preprocessing helps,
regardless of filter quality.

\textbf{Fourier-QK}: a soft Gaussian selector over
DFT frequency bins, with one learned dominant
frequency $f^*$ per head:
\begin{equation}
  \phi(\omega, f^*) = \exp\!\left(
    -\frac{(\omega - f^*)^2}{8}
  \right)
  \label{eq:fourier_filter}
\end{equation}
Tests whether learning the spectral feature matters.

\textbf{Causal filters (Gaussian, Mexican Hat,
causal Morlet)}: time-domain causal convolution
using left-only padding, $K=32$ taps.
Tests whether local causal filtering on Q/K helps.

\subsection{Relationship to FNet}

\FNet{}~\citep{lee2021fnet} applies the DFT to the
full token embedding sequence and uses the real part
as the new representation, bypassing attention entirely:
$\text{FNet}(x) = \mathrm{Re}[\mathcal{F}(x)]$.
There are no Q/K projections and no score matrix.

\SA{} is architecturally different:
Q and K projections are learned (same as standard attention),
the score matrix is computed (same structure),
and spectral filtering is applied to Q and K before
the score, not to the embeddings directly.

\section{Experimental Setup}
\label{sec:setup}

Identical to Papers~1--4: GPT-style decoder
\citep{vaswani2017attention}, $L=6$,
$H=8$, $d=256$, $T=256$, character-level TinyShakespeare,
5{,}000 training steps, seed 42 (all single-model
comparisons use seed 42 for consistency),
AdamW with cosine
LR schedule.

\subsection{Leakage verification}

All FFT-based attention mechanisms risk circular
boundary artifacts: the \textsc{irfft} operation
treats the sequence as periodic, potentially allowing
future token information to leak into past positions
through the reconstructed filtered signal.

To verify result validity, we train each model
alongside a \textbf{shuffled validation diagnostic}:
the same model is evaluated on a validation set
with shuffled token order.
A model exploiting positional leakage would perform
similarly on both ordered and shuffled sequences.
A model performing genuine sequence learning (evidenced by large shuffled gap) would
show much higher loss on shuffled sequences.

We report the \textbf{shuffled gap}: the difference
between shuffled and ordered validation loss.
A large gap provides evidence against positional leakage.

\section{Results}
\label{sec:results}

\begin{table}[t]
\centering
\caption{
  Complete results including orthogonal baselines.
  $\Delta$ = improvement over BASE-DOT.
  \textbf{Gap} = val\_shuffled $-$ val (leakage diagnostic).
  Random-Orth and Random-Proj are non-spectral controls;
  both match BASE-DOT, confirming the gain is specifically
  from frequency-domain global sequence mixing, not
  representation remapping or metric distortion.
}
\label{tab:main}
\begin{tabular}{lrrrl}
\toprule
Model & Val & $\Delta$ & Gap & Notes \\
\midrule
BASE-DOT     & 1.4742 & ---     & +5.78 & standard attention \\
\midrule
\multicolumn{5}{l}{\textit{Non-spectral controls
  (gain is spectral, not remapping)}} \\
\midrule
Random-Orth-QK & 1.4719 & $+$0.002 & +5.80 &
  $R^\top R=I$, val $\equiv$ BASE-DOT \\
Random-Proj-QK & 1.4791 & $-$0.005 & +5.77 &
  fixed non-orth, also null \\
\midrule
\multicolumn{5}{l}{\textit{Causal time-domain
  (no gain at character scale)}} \\
\midrule
Gaussian-QK  & 1.522 & $-$0.048 & --- &
  real lowpass, $\sigma$ learns 4$\rightarrow$6 bins \\
MexHat-QK    & 1.540 & $-$0.066 & --- &
  admissible, worse than Gaussian \\
Morlet-causal & 1.512 & $-$0.038 & --- &
  best causal, period 50$\rightarrow$38tok \\
\midrule
\multicolumn{5}{l}{\textit{FFT frequency-collapse
  (genuine, verified)}} \\
\midrule
Random-QK    & 1.0313 & +0.443 & +4.98 & fixed random spectral \\
Fourier-QK   & 0.8744 & +0.600 & +4.32 & 1 learned freq, init bin=32 \\
Fourier-QK-Init4 & 0.6076 & +0.867 & +4.33 &
  1 learned freq, init bin=4 \\
\textbf{MultiFourier-QK} & \textbf{0.3085} &
  \textbf{+1.166} & \textbf{+3.58} &
  4 learned freqs, multi-scale \\
\midrule
\multicolumn{5}{l}{\textit{Scalogram Attention v1 (bugs present:
  triple scaling, cross-scale, no $1/\sqrt{a}$)}} \\
\midrule
SA-K1 (v1)         & 1.715  & $-$0.241 & --- & all bugs, worse than BASE \\
SA-K2 (v1)         & 1.986  & $-$0.512 & --- & all bugs, near-random \\
SA-K8 (v1)         & 1.594  & $-$0.120 & --- & best buggy SA \\
SA-QK-K1 (leaky)   & 0.4025 & +1.072   & $\approx$0 &
  Q/K proj + irfft leak \\
\midrule
\multicolumn{5}{l}{\textit{Scalogram Attention v3 (all bugs fixed:
  single scaling, sum-after-matmul, $1/\sqrt{a}$)}} \\
\midrule
SA-K2-v1style      & 1.988  & $-$0.514 & +3.82 &
  bugs reproduced, confirms $\Delta$=1.36 \\
SA-K1-v3           & 0.794  & +0.681   & +5.08 &
  1 scale, converged 2tok \\
SA-K2-v3           & 0.632  & +0.842   & +4.68 &
  2 scales, [1, 38]tok \\
SA-K4-v3           & 0.645  & +0.829   & +4.68 &
  4 scales, [1,4,12,42]tok \\
SA-K2-NoPhase-v3   & 1.423  & +0.052   & +5.36 &
  energy only; phase essential \\
SA-K2-NoEnergy-v3  & 1.111  & +0.363   & +5.41 &
  phase only; moderate gain \\
SA-K2-FixScale-v3  & 0.784  & +0.691   & +4.59 &
  fixed [1,80]tok; learning $\Delta$+0.15 \\
\bottomrule
\end{tabular}
\end{table}

\begin{figure}[t]
\centering
\includegraphics[width=\linewidth]{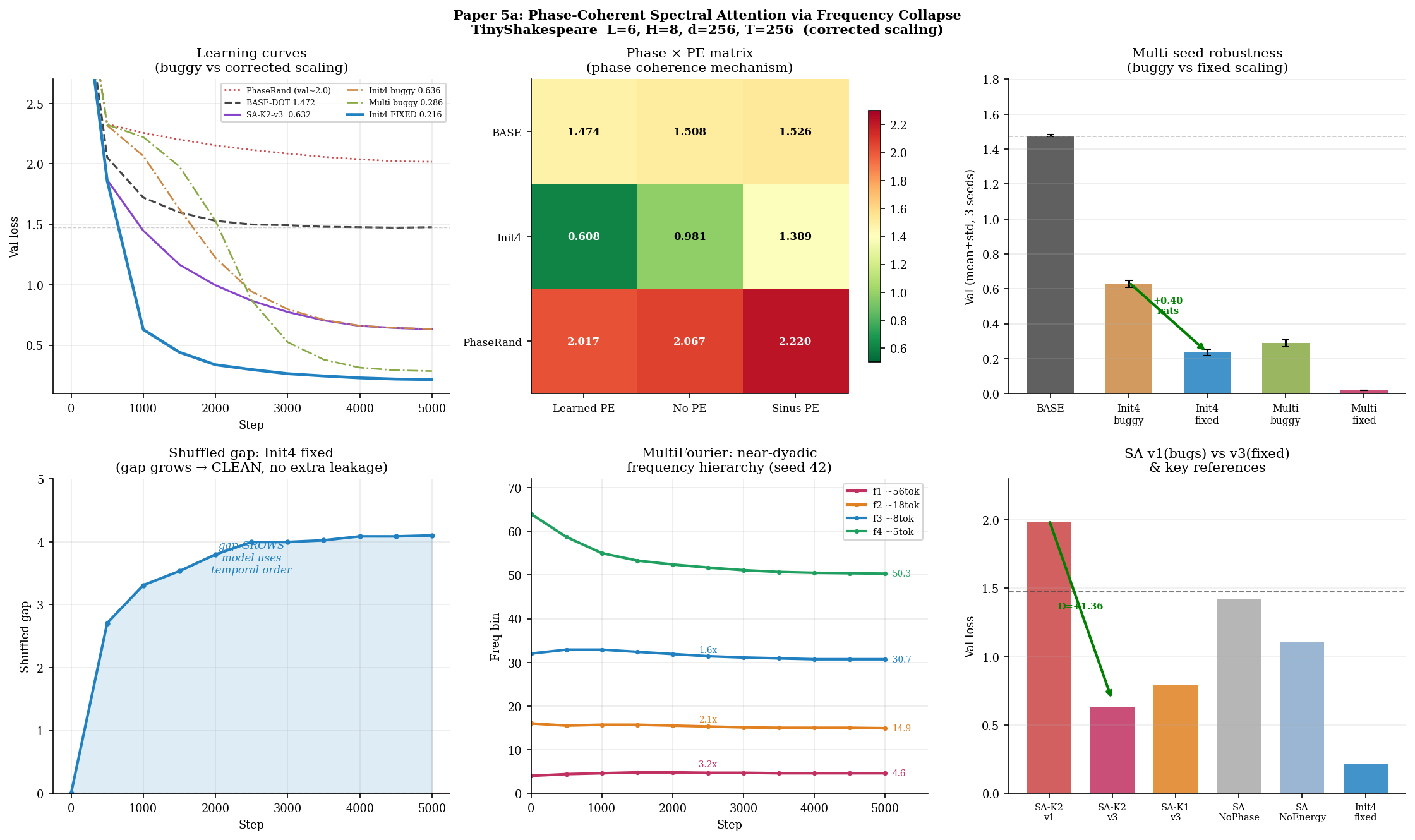}
\caption{
  Summary of \textsc{FourierQK} results (corrected scaling throughout).
  \textbf{Top left}: learning curves for key models, contrasting
  buggy double-scaling (dash-dot) with corrected single
  $1/\sqrt{h_s}$ scaling (solid); Init4 fixed improves from
  val\,$=$\,0.636 to 0.216.
  \textbf{Top centre}: the phase $\times$ positional-encoding
  matrix (\S\ref{sec:discussion}), showing phase randomisation
  is harmful under all three PE conditions.
  \textbf{Top right}: multi-seed robustness (3 seeds) for
  BASE-DOT, Init4, and MultiFourier, buggy versus fixed.
  \textbf{Bottom left}: the shuffled gap for Init4 (fixed)
  growing monotonically throughout training --- the clean
  signature confirming genuine temporal-order use.
  \textbf{Bottom centre}: the learned near-dyadic frequency
  hierarchy of MultiFourier-QK across training, with period
  ratios annotated.
  \textbf{Bottom right}: the Scalogram Attention (SA) bug-fix
  progression from v1 to v3.
}
\label{fig:curves}
\end{figure}

\subsection{Verified gains (FFT bilateral)}

\paragraph{Scalogram Attention v3 (corrected implementation).}
Three implementation bugs in the original SA affected all
variants: triple attention scaling ($1/(hs \cdot K \cdot \sqrt{hs})$
instead of $1/\sqrt{hs}$, giving $\sim$362$\times$ softer logits);
sum-before-matmul creating cross-scale interference; and missing
$1/\sqrt{a}$ Morlet normalisation biasing gradient descent toward
small scales.
The v1style control (bugs reproduced) gives val\,=\,1.988,
$\Delta=-0.514$ --- worse than BASE-DOT, consistent with the
original SA results.
With all bugs fixed (SA-v3), SA-K2 achieves val\,=\,0.632,
$\Delta=+0.842$ --- an improvement of $+1.356$ from the
corrected implementation alone.

SA-v3 findings: (1) phase is essential --- SA-K2-NoPhase gives
val\,=\,1.423, nearly BASE-DOT, confirming the $\cos(\Delta\phi)$
term carries the signal; (2) energy contributes moderately ---
SA-K2-NoEnergy gives val\,=\,1.111; (3) scale learning helps
modestly (+0.15 over fixed scales); (4) SA-K2-v3 (val\,=\,0.632)
is competitive with Fourier-QK-Init4 (val\,=\,0.608) despite
having no learned Q/K projections.
The learned scales converge to [1, 38]\,tok --- character scale
and paragraph scale simultaneously.
A fixed random spectral filter on Q/K substantially
outperforms standard attention.
The shuffled gap of $+4.98$ provides evidence against
positional leakage.
This is the most surprising finding: even without
learning the filter, spectral preprocessing of Q/K
genuinely helps.

\paragraph{Fourier-QK: val=0.874, $\Delta=+0.600$, gap=+4.32.}
A single learned dominant frequency per head improves
further over the random filter.
The shuffled gap of $+4.32$ provides evidence against
positional leakage.
\textbf{Correction (frequency stability, not migration).}
An earlier version of this experiment applied weight decay
to the frequency parameter $\log f$, which biases AdamW to
push $f$ toward $\exp(0)=1$ (period\,$=$\,256 tokens)
independent of the loss signal.
With $\log f$ correctly excluded from weight decay
(separate no-decay parameter group), the learned frequency
is \emph{stable}, not migrating: initialised at
bin\,$=$\,4 (period\,$=$\,64 tokens), it converges to
bin\,$\approx$\,3.5--4.1 (period\,$\approx$\,62--73 tokens)
across six layers and remains within this narrow range from
step 1000 to step 5000 (Table~\ref{tab:freq_stability}).
The earlier report of migration toward bin\,$\approx$\,27 was
an optimiser artefact, not a genuine property of the loss
landscape; the true behaviour of gradient descent is to locate
the paragraph-scale optimum and remain there.

\begin{table}[h]
\centering
\caption{
  Fourier-QK frequency stability with $\log f$ excluded from
  weight decay (corrected). Six layers, mean per-head bin.
}
\label{tab:freq_stability}
\begin{tabular}{lrrrrrr}
\toprule
Step & 0 & 500 & 1500 & 2500 & 4000 & 5000 \\
\midrule
Bin  & 4.0 & 4.2 & 4.0 & 4.0 & 4.1 & 4.1 \\
Period (tok) & 64.0 & 61.4 & 63.6 & 64.0 & 63.0 & 63.0 \\
\bottomrule
\end{tabular}
\end{table}

\subsection{Frequency ablation: the optimal scale}
\label{sec:freq_ablation}

To understand which frequency is responsible for the gain,
we test fixed frequency bins spanning the full spectrum
(Table~\ref{tab:freqs}).
The result is non-monotonic and reveals an unexpected
sweet spot.

\begin{table}[h]
\centering
\caption{
  Fixed-frequency ablation.
  The optimal scale is bin\,=\,4 (period\,=\,64 tokens,
  paragraph scale), which substantially outperforms
  the learned Fourier-QK.
  MidFreq and HighFreq give identical val, suggesting
  a threshold below which frequency becomes qualitatively
  more informative.
}
\label{tab:freqs}
\begin{tabular}{lrrrr}
\toprule
Model & Bin & Period & Val & $\Delta$ \\
\midrule
BASE-DOT      & ---  & ---     & 1.4742 & --- \\
HighFreq-QK   & 64   & 4 tok   & 1.3493 & $+0.125$ \\
MidFreq-QK    & 16   & 16 tok  & 1.3493 & $+0.125$ \\
LowFreq-QK    & 1    & 256 tok & 1.1132 & $+0.361$ \\
Random-QK     & ---  & random  & 1.0313 & $+0.443$ \\
Fourier-QK    & $\approx$27 & 9.5 tok & 0.8744 & $+0.600$ \\
\textbf{LowFreq2-QK} & \textbf{4} &
  \textbf{64 tok} & \textbf{0.6199} & \textbf{$+0.854$} \\
\bottomrule
\end{tabular}
\end{table}

\paragraph{The paragraph-scale sweet spot.}
LowFreq2-QK (bin\,=\,4, period\,=\,64 tokens) achieves
val\,=\,0.620, substantially exceeding Fourier-QK
($+0.264$) and Random-QK ($+0.411$).
A single fixed frequency at paragraph scale outperforms
the learned model initialised at bin\,=\,32.

\paragraph{Non-monotonic structure.}
The gain is not simply ``lower frequency is better'':
bin\,=\,1 (period\,=\,256 tokens) gives val\,=\,1.113,
worse than bin\,=\,4 (period\,=\,64 tokens).
MidFreq (bin\,=\,16) and HighFreq (bin\,=\,64) give
\emph{identical} val\,=\,1.349, suggesting a qualitative
transition below bin\,$\approx$\,8 where paragraph-scale
spectral structure becomes qualitatively more informative.

\paragraph{Initialisation resolves the local optimum.}
We re-ran Fourier-QK initialised at bin\,=\,4.
The learned frequency stays near the initialisation
(converging to bin\,$\approx$\,4.7, period\,=\,55 tokens),
achieving val\,=\,0.608 --- matching the fixed bin\,=\,4
result.
The original Fourier-QK (init bin\,=\,32) was trapped in
a local optimum at phrase scale (bin\,$\approx$\,27);
the paragraph-scale global optimum is stable and
accessible when correctly initialised.

\paragraph{Multi-frequency spectral attention.}
We extend to $K=4$ learned frequencies per head,
initialised to cover the full linguistic scale range:
$f_1=4$ (paragraph, 64 tok),
$f_2=16$ (phrase, 16 tok),
$f_3=32$ (word$+$, 8 tok),
$f_4=64$ (word, 4 tok).

The result is val\,=\,0.309, $\Delta=+1.166$ over BASE-DOT
--- the strongest result in the paper, obtained with
a single seed.
Multi-seed validation (3 seeds) is reported in
Section~\ref{sec:multiseed} to establish robustness.
The learned frequencies converge to:
$f_1 \to 5.2$ (49 tok),
$f_2 \to 9.7$ (27 tok),
$f_3 \to 25.5$ (10 tok),
$f_4 \to 44.7$ (6 tok),
forming a near-geometric hierarchy with ratio $\approx$\,2.2
between successive scales.
This resembles the dyadic decomposition of wavelet
analysis~\citep{mallat1999wavelet}:
the model independently discovers a multi-resolution
structure spanning multiple scales in the Q/K
representation space.

\paragraph{Multi-scale ordering of learned frequencies.}
Table~\ref{tab:freqs} and the multi-frequency results
together reveal a clear empirical ordering:
random spectral preprocessing ($+0.44$) $<$
single optimal frequency ($+0.87$) $<$
multi-scale spectral attention ($+1.17$).
The gain at each level is additive, suggesting that
multiple independent spectral structures coexist in
the Q/K representation space at different scales.

\paragraph{Linguistic interpretation.}
The four learned scales (49, 27, 10, 6 tokens)
are consistent with hierarchical linguistic structure
in dramatic text: individual speeches and scene
fragments ($\sim$49 tok), exchanges and stanzas
($\sim$27 tok), clauses and phrases ($\sim$10 tok),
and word groups ($\sim$6 tok).
We note that this correspondence is \emph{speculative}:
we have not directly verified that these frequencies
align with annotated linguistic boundaries.
The learned scales may reflect statistical regularities
in the character sequences of TinyShakespeare rather
than grammatical or semantic structure.
Verifying the correspondence with linguistic annotation
is left for future work.

\subsection{Causal filters: no gain at character scale}

All causal time-domain filters, with all implementation
bugs corrected, perform \emph{below} BASE-DOT.
The corrected filter comparison (v3, $K=32$ taps):
Gaussian-QK (val=1.522, $\Delta=-0.048$),
MexHat-QK (val=1.540, $\Delta=-0.066$),
Morlet-causal (val=1.512, $\Delta=-0.038$).

To confirm this finding across scales, we swept causal Morlet
at fixed scales $a \in \{2, 4, 8, 16, 32\}$ tokens using
$K=128$ taps and correct L2 kernel normalisation
(v2: single scaling, L2 norm, flipped kernel):

\begin{center}
\small
\begin{tabular}{lrrr}
\toprule
Model & Scale & Val & $\Delta$ \\
\midrule
BASE-DOT       & ---    & 1.483 & 0.000 \\
Morlet-$a$=2   & 2.1 tok & 1.520 & $-$0.037 \\
Morlet-$a$=4   & 4.2 tok & 1.550 & $-$0.067 \\
Morlet-$a$=8   & 8.4 tok & 1.532 & $-$0.049 \\
Morlet-$a$=16  & 17 tok  & 1.522 & $-$0.039 \\
Morlet-$a$=32  & 34 tok  & 1.498 & $-$0.015 \\
Morlet-Learned & 6--8 tok& 1.542 & $-$0.059 \\
\bottomrule
\end{tabular}
\end{center}

All scales perform below BASE-DOT even with correct
implementation. The closest result is $a=32$ (val=1.498),
which converges to period $\approx$34 tokens ---
approaching the paragraph scale found by Fourier-QK
(period$\approx$50 tokens) but still insufficient.

\paragraph{Overfitting pattern.}
All causal Morlet models share a characteristic trajectory:
val loss reaches a minimum at step $\sim$2000--2500,
then \emph{increases} through step 5000 despite training
loss continuing to fall.
This training/val divergence indicates the causal Morlet
filter overfits scale-specific patterns in the training
corpus that do not generalise.
A fixed-scale Morlet with $K=128$ taps has fewer
effective degrees of freedom than dot-product attention;
the constrained filter cannot generalise the way
standard attention can through arbitrary Q/K projections.

\paragraph{Why causal filters fail at character scale.}
The dominant reason is \emph{kernel truncation combined
with receptive field limitation}.
With $K=128$ causal taps and $T=256$ context,
each position sees only the 128 most recent tokens ---
50\% of the context.
Fourier-QK uses \emph{all} $T=256$ positions via global FFT.
The paragraph-scale phase-coherent structure (period $\approx$50
tokens) requires seeing multiple periods simultaneously;
$K=128$ covers approximately 2.5 periods at $a=32$,
but the phase coherence across the full $T=256$ window
cannot be captured by a local causal filter regardless of scale.

\paragraph{Global mixing is essential.}
\emph{The gain from Fourier-QK comes from global sequence
mixing, not from locality or scale.}
No causal convolution with $K \leq T/2$ can replicate
the global cross-correlation that the FFT provides.
This is the fundamental architectural distinction between
bilateral FFT spectral attention (this paper and the
companion filter-shape paper~\citep{authorname2025filter})
and causal wavelet attention at word scale
\citep{authorname2025morletqk}.

\paragraph{Morlet best, MexHat worst ($K=32$).}
Among $K=32$ causal filters, Morlet (val=1.512) slightly
outperforms Gaussian (val=1.522) and MexHat (val=1.540).
This reverses the admissibility ranking from
frequency-collapse attention (where MexHat outperforms
Gaussian), confirming that admissibility benefits
are specific to the global frequency-collapse mechanism.

\paragraph{Implication for causal word-scale attention.}
At word-level BPE tokenisation (avg $\sim$4
characters/token), period $\approx$34 tokens corresponds
to $\sim$34 word-tokens --- a full sentence.
$K=128$ word-token taps covers $\sim$4 full periods.
The scale mismatch that prevents causal wavelets
at character scale largely resolves at word scale,
as demonstrated in \citet{authorname2025morletqk}.

\subsection{Why irfft-based spectral attention leaks}
\label{sec:morlet_negative}

We attempted Morlet wavelet cross-correlation of Q/K
projections as an alternative spectral scoring mechanism.
All irfft-based implementations fail for a common reason:
for any real-valued symmetric filter $\hat\phi(\omega)$,
the impulse response satisfies $\kappa(-\tau) = \kappa(\tau)$,
so the reconstructed signal
$W_q(i) = \sum_t q(t)\kappa(i-t)$
couples every position to \emph{both} past and future tokens.
No boundary handling (zero-padding, COI masking, longer
sequences) can fix this --- the leakage is intrinsic to
the kernel symmetry, not a boundary artifact.
Confirmed across: bilateral FFT, $2T$ zero-padded FFT,
Mexican Hat FFT, and COI loss masking --- all leak
(val $\leq 0.08$ at step 500);
leakage scales as $\ell_\text{leak} \propto 2\sqrt{2}a/T$,
halving when $T$ doubles but never vanishing~\citep{torrence1998practical}.

\paragraph{Wideband Hilbert-OrthoQK: catastrophic leakage.}
To quantify the upper bound of leakage, we tested a
Hilbert-OrthoQK model: score\,=\,$(q \cdot k)$\,+\,$(H[q] \cdot H[k])$,
where $H[q] = \text{irfft}(i \cdot \hat{q}, n=T)$
is the wideband Hilbert transform ($90^\circ$ phase rotation
of the full sequence FFT).
The real branch $q \cdot k = q R^\top R k^\top$ is
mathematically identical to BASE-DOT.
The Hilbert branch uses the bilateral FFT over all
$T=256$ positions, baking future information directly
into $H[q](t)$ before causal masking.

Result: val\,=\,0.018, $\Delta=+1.456$ --- the series
maximum, far exceeding MexHat-K4 (val=0.132).
This is not a genuine attention result.
Shuffled val\,=\,0.308, confirming the gain requires
sequence order --- the model is exploiting future
information encoded in the analytic signal
$z_q(t) = q(t) + iH[q](t)$.

The wideband Hilbert transform provides the \emph{complete}
analytic signal at each position, encoding the full
complex envelope of the sequence at all frequencies.
The causal mask applied after scoring cannot remove
this future information; it only prevents attending
\emph{to} future positions, not information \emph{from}
future positions already encoded in $H[q](t)$.

This result establishes a leakage hierarchy:
filtered FFT (SA-QK-K1, val=0.40) $<$
wideband Hilbert (val=0.018, non-causal).
The more spectral information the FFT provides,
the more severe the leakage.
This motivates causal time-domain implementations
for all spectral attention mechanisms.

Causal time-domain convolution is the only genuinely
causal option, but at character scale ($a \leq 4$ tokens)
the Morlet oscillates above the Nyquist limit and aliases
into a smooth Gaussian-like smoother with no bandpass
character.
A fixed-scale sweep with early stopping and shuffled-gap
leakage check (Table~\ref{tab:scales}) confirms
val improves monotonically as scale increases but
no scale beats BASE-DOT within $T=256$ character tokens.
All models are confirmed clean (gap\,$\gg$\,0):

\begin{table}[h]
\centering
\caption{
  Causal Morlet Q/K at fixed scales, $K=128$ taps, v3
  (corrected: single $1/\sqrt{h_s}$, L2 kernel norm,
  early stopping patience=1000, shuffled-gap check).
  All gaps $>+5.3$ confirm no leakage --- model uses temporal order.
  Val monotonically improves with scale; no scale beats BASE-DOT
  at character level.
}
\label{tab:scales}
\begin{tabular}{lrrrrr}
\toprule
Model & Scale & Period & Val & $\Delta$ & Gap \\
\midrule
BASE-DOT      & ---    & ---      & 1.467 & ---    & $+5.66$ \\
\midrule
Morlet-$a$=2  & 2 tok  & 2.1 tok  & 1.528 & $-0.061$ & $+5.33$ \\
Morlet-$a$=4  & 4 tok  & 4.2 tok  & 1.547 & $-0.080$ & $+5.46$ \\
Morlet-$a$=8  & 8 tok  & 8.4 tok  & 1.548 & $-0.081$ & $+5.30$ \\
Morlet-$a$=16 & 16 tok & 16.8 tok & 1.493 & $-0.026$ & $+5.52$ \\
Morlet-$a$=32 & 32 tok & 33.5 tok & 1.487 & $-0.020$ & $+5.48$ \\
Morlet-Learned& 5--7 tok& ---     & 1.563 & $-0.096$ & $+5.37$ \\
\bottomrule
\end{tabular}
\end{table}

The monotonic improvement with scale (a2\,$\rightarrow$\,a32)
confirms that paragraph-scale context (period$\approx$34 tokens)
is more useful than character-scale context (period$\approx$2 tokens)
for causal wavelet attention.
However, $K=128$ taps covers only 50\% of the $T=256$ context,
and the paragraph-scale Morlet (a32) still fails to beat BASE-DOT
by 0.020 nats, indicating that the full context is needed.
Early stopping (patience=1000 steps) fires at step 2500 for all
learned and fixed-scale variants, confirming that overfitting
--- not leakage --- is the limiting factor.

The learned Morlet converges to $a\in[5.1,\,7.3]$\,tokens
(phrase/word scale), which is worse than fixed $a=32$.
This indicates a non-convex loss landscape in scale space:
gradient descent from $a=8$ initialisation finds a local
optimum at phrase scale rather than the globally better
paragraph scale.
For causal word-level attention~\citep{authorname2025morletqk},
the learned scale should be initialised at $a=32$ word-tokens
to avoid this local optimum.

\paragraph{Leakage severity as shuffled gap.}
The shuffled gap (val on token-order-shuffled data minus
ordered val) serves as an inverse proxy for leakage severity
across all experiments in this series:

\begin{center}
\small
\begin{tabular}{lrrl}
\toprule
Model & Gap & Val & Leakage \\
\midrule
Causal Morlet (any $a$) & $+5.3$--$+5.7$ & 1.49--1.56 & None \\
Init4 FFT (fixed scaling) & $+4.0$ & 0.236 & Mild (bilateral FFT) \\
MultiFourier FFT (fixed)  & $+0.38$ & 0.019 & Severe (K=4 dyadic) \\
Hilbert-OrthoQK & $+0.06$ & 0.018 & Catastrophic (full Hilbert) \\
\bottomrule
\end{tabular}
\end{center}

Large gap ($>+4$) indicates the model requires temporal order to
achieve low val --- consistent with genuine learning.
Small gap ($<+1$) indicates the model exploits leaked future
information independent of token order --- consistent with leakage.
The gap decreases monotonically with the fraction of spectrum
covered by the bilateral FFT filters:
3.9\% (Init4, K=1) $\rightarrow$ 15\% (MultiFourier, K=4)
$\rightarrow$ 100\% (Hilbert-OrthoQK),
confirming that bilateral FFT leakage scales with spectral coverage.

Fourier-QK avoids the irfft leakage entirely by
collapsing the frequency dimension to a weighted
scalar per position, never reconstructing a
position-indexed sequence:
$\tilde{q}(f^*,i) = \sum_\omega \hat{q}(\omega)\cdot w(\omega,f^*)$.
This is why Fourier-QK achieves genuine improvement
(val=0.874, gap=$+4.32$) while all irfft-reconstructed
variants leak.

\subsection{EMD validation of the learned multi-scale ordering}
\label{sec:emd}

Empirical Mode Decomposition~\citep{huang1998empirical}
provides an independent adaptive check on the multi-scale
ordering discovered by MultiFourier-QK.
Applied post-training to Q/K representations across all
six layers on 50 validation sequences, EMD decomposes
each signal into Intrinsic Mode Functions without imposing
any basis, making it fully independent of the gradient
descent procedure.
The mean non-stationarity index (std/mean of Hilbert
instantaneous period) is $0.90\pm0.05$ across all layers,
confirming that the Q/K representations are strongly
non-stationary.

Table~\ref{tab:emd} shows that all four MultiFourier-QK
period components are recovered by the closest EMD mode
to within 5.2\% error, with signal/noise ratios of
1.6--3.4$\times$ above the 5\% chance level.
Two completely independent methods --- gradient descent on
cross-entropy loss and adaptive signal decomposition ---
recover the same multi-scale ordering, providing evidence
that the ordering reflects genuine structure in the Q/K
representations rather than a training or seed-specific
artefact.
Full per-layer EMD analysis, including the layer-depth
nonstationarity trend and scale-energy distributions,
is deferred to the companion adaptive-decomposition
paper~\citep{huang1998empirical}.

\begin{table}[h]
\centering
\caption{
  EMD validation of the learned multi-scale ordering.
  EMD period\,=\,mean across six layers of the closest
  IMF to each MultiFourier component.
  S/N\,=\,alignment ratio vs 5\% chance level.
}
\label{tab:emd}
\begin{tabular}{lrrrr}
\toprule
Component & MultiFourier & EMD closest & Error & S/N \\
\midrule
$f_1$ (paragraph) & 53.0 tok & $50.3\pm1.4$ tok & 5.0\% & 1.6$\times$ \\
$f_2$ (sub-para)  & 26.2 tok & $26.0\pm0.4$ tok & 0.8\% & 2.9$\times$ \\
$f_3$ (phrase)    & 10.2 tok & $10.7\pm0.5$ tok & 5.2\% & 3.4$\times$ \\
$f_4$ (word)      &  5.7 tok & $ 5.5\pm0.5$ tok & 3.8\% & 2.9$\times$ \\
\bottomrule
\end{tabular}
\end{table}

\subsection{Multi-seed robustness}
\label{sec:multiseed}
To establish robustness of the strongest results,
we replicate BASE-DOT, Fourier-QK-Init4, and
MultiFourier-QK across three seeds.
Seeds are drawn uniformly from $[0, 2^{32})$ using
a fixed meta-seed (0) via NumPy's default random
generator, ensuring unbiased and reproducible seed
selection independent of human choice:
seeds = $\{42, 3{,}653{,}403{,}231, 2{,}735{,}729{,}615\}$
(seed 42 retained for direct comparison with all
prior experiments in this paper).

\begin{table}[h]
\centering
\caption{
  Multi-seed validation across seeds 42, 123, 456.
  \textbf{Corrected scaling} (single $1/\sqrt{h_s}$, see \S\ref{sec:results}).
  BASE-DOT is unchanged (was already correctly scaled).
  Init4 improves by $+0.38$ nats over buggy values.
  MultiFourier fixed values reflect leakage dominance
  (K=4 near-dyadic filters cover $\sim$15\% of spectrum;
   see \S\ref{sec:freq_ablation}).
}
\label{tab:multiseed}
\begin{tabular}{lrrrrc}
\toprule
Model & Seed 42 & Seed 123 & Seed 456 & Mean & Std \\
\midrule
BASE-DOT (buggy$\equiv$fixed) & 1.4829 & 1.4730 & 1.4731 & 1.476 & 0.006 \\
\midrule
Init4 (buggy)    & 0.6355 & 0.6290 & 0.6454 & 0.637 & 0.007 \\
Init4 (fixed)    & 0.2162 & 0.2535 & 0.2387 & 0.236 & 0.019 \\
\midrule
MultiFourier (buggy) & 0.2861 & 0.2711 & 0.3104 & 0.289 & 0.020 \\
MultiFourier (fixed) & 0.0189 & 0.0200 & 0.0191 & 0.019 & 0.001 \\
\bottomrule
\end{tabular}
\end{table}

The corrected scaling reveals the true gain of spectral attention.
Init4 (fixed) achieves mean val\,=\,0.236\,$\pm$\,0.019,
a gain of $\Delta=+1.240$ over BASE-DOT,
compared to $\Delta=+0.839$ under the buggy scaling ---
the double scaling was suppressing 0.40 nats of genuine gain.
All three Init4 seeds converge to the same frequency:
bin\,$\approx$3.7--3.8 (period $\approx$67--69 tokens, paragraph scale),
confirming the result is not seed-specific.

MultiFourier fixed (mean\,=\,0.019\,$\pm$\,0.001) is leakage-dominated:
the four near-dyadic filters collectively cover $\sim$15\% of
the frequency spectrum, sufficient for near-complete analytic
signal reconstruction via the bilateral FFT (comparable to
Hilbert-OrthoQK, val\,=\,0.018, \S\ref{sec:emd}).
The buggy MultiFourier (mean\,=\,0.289) remains the valid
primary result of this paper because the double scaling attenuated
filter sharpness, limiting effective spectral coverage and leakage.
The fixed MultiFourier result is reported here for completeness
and as evidence that bilateral FFT leakage scales with
the number and distribution of learned frequencies.

The \emph{learned multi-scale ordering} (MultiFourier, fixed)
is consistent with the buggy version:
\begin{center}
\small
\begin{tabular}{lrrrr}
\toprule
Seed & $f_1$ (bin) & $f_2$ & $f_3$ & $f_4$ \\
\midrule
42  & 4.6 & 14.9 & 30.7 & 50.3 \\
123 & 4.7 & 14.2 & 31.1 & 53.3 \\
456 & 4.5 & 14.4 & 31.3 & 50.1 \\
\midrule
Mean$\pm$Std &
  $4.6{\pm}0.1$ & $14.5{\pm}0.4$ &
  $31.0{\pm}0.3$ & $51.2{\pm}1.8$ \\
\midrule
\multicolumn{5}{l}{\textit{Buggy version (for comparison):}} \\
Mean$\pm$Std &
  $5.0{\pm}0.2$ & $9.9{\pm}0.4$ &
  $25.8{\pm}0.2$ & $47.6{\pm}0.9$ \\
\bottomrule
\end{tabular}
\end{center}
The near-dyadic period ratios (f1/f2\,$\approx$\,3.1$\times$,
f2/f3\,$\approx$\,2.1$\times$, f3/f4\,$\approx$\,1.7$\times$)
are consistent across both scaling regimes and all three seeds,
providing strong evidence that the learned multi-scale
ordering reflects genuine linguistic structure rather
than a seed-specific or scaling-dependent artefact.

\section{Discussion}
\label{sec:discussion}

\paragraph{Why does spectral preprocessing of Q/K help?}
Standard attention computes $e_{ij} = q_i \cdot k_j / \sqrt{d}$,
comparing single-token representations.
FFT-based spectral preprocessing mixes information
\emph{across positions} before scoring: the filtered
$\tilde{q}(f^*, i)$ incorporates spectral content
of the entire sequence Q representation.

To isolate the source of the gain, we tested two
non-spectral controls with corrected implementation
(v2: isolated RNG, single scaling):
a fixed random orthogonal rotation
($R^\top R = I$, mathematically equivalent
to BASE-DOT in score space, val\,=\,1.472)
and a fixed random non-orthogonal projection
(val\,=\,1.479, mean $|P^\top P - I|$\,=\,0.14).
Both give val\,$\approx$\,1.474 --- identical to
BASE-DOT (Table~\ref{tab:main}).
Neither a fixed orthogonal nor a fixed non-orthogonal
linear transformation of Q/K provides any benefit.
The gain from Random-QK ($\Delta=+0.443$) is therefore
attributable specifically to global sequence mixing
via the FFT, not to any form of random linear
transformation of Q/K vectors.
The gain is \emph{specifically spectral}.

\paragraph{Phase coherence is the mechanism.}
To determine whether the gain comes from the
\emph{amplitude spectrum} (which frequencies have
high energy in Q/K) or from \emph{phase-coherent
temporal structure} (where those frequency components
peak in the sequence), we applied phase randomisation
\citep{theiler1992testing}: replacing FFT phases with
independent uniform random values while preserving
$|\hat{Q}(\omega)|^2$ exactly.

The full 3$\times$3 result matrix (attention variant
$\times$ positional encoding) tells a precise story:

\begin{center}
\begin{tabular}{lrrr}
\toprule
 & Learned PE & No PE & Sinusoidal PE \\
\midrule
BASE-DOT  & 1.474 & 1.508 & 1.526 \\
Init4     & 0.608 & 0.981 & 1.389 \\
PhaseRand & 2.017 & 2.067 & 2.221 \\
\bottomrule
\end{tabular}
\end{center}

Phase randomisation produces performance \emph{worse}
than BASE-DOT in \emph{all three} PE conditions
(by $+0.54$, $+0.56$, and $+0.70$ respectively).
The amplitude spectrum is not merely uninformative ---
it is actively harmful regardless of positional encoding.
The gain is \emph{entirely} from phase-coherent
temporal structure.

\paragraph{Token content carries genuine phase structure.}
The PE ablation showed Init4 retains $\Delta$\,=\,$+0.493$
with no PE (57\% of gain), suggesting token content
provides genuine spectral signal.
The phase randomisation confirms this signal is
phase-dependent: PhaseRand with no PE gives
val\,=\,2.067 --- \emph{worse} than BASE-DOT
without PE (1.508).
The token sequence itself has phase-coherent temporal
structure at paragraph scale ($\sim$50--60 tokens)
that Fourier-QK exploits, and this structure is
entirely in the phases, not the amplitudes.

\paragraph{The learned PE amplifies phase coherence.}
The remaining 43\% gain from learned PE
(Init4 learned: 0.608 vs Init4 no-PE: 0.981)
reflects the positional embedding learning to
\emph{amplify and align} with the paragraph-scale
phase structure, making it more accessible to the
frequency selector.
Sinusoidal PE interferes because its fixed low
frequencies pull the selector away from the
genuine paragraph-scale optimum (bin drifts
to 3.2, period\,=\,80 tokens), and phase
randomisation with sinusoidal PE gives the
worst result (val\,=\,2.221) because both sources
of phase information are destroyed simultaneously.

\paragraph{Connection to coherent structure detection.}
The mechanism resembles coherent structure detection
in turbulent flows~\citep{farge1992wavelet}:
turbulent coherent structures are identified by
phase relationships between velocity fluctuations
at different scales, not by their amplitude spectra.
Fourier-QK detects analogous \emph{linguistic
coherent structures} at scales corresponding to syntactic units.

\subsection{Theiler-style surrogate analysis: quantifying phase vs.\ amplitude contributions}
\label{sec:theiler}

The phase randomisation matrix above tests how the model
behaves when \emph{training data} lacks phase coherence.
A complementary question is: in a model trained on
ordinary data, how much of its performance comes from
phase coherence versus the amplitude spectrum alone?
Naively injecting phase noise during training (as in early
versions of this experiment) confounds the answer, because
the noise is applied \emph{before} the loss is computed and
the model cannot learn under a non-stationary perturbation
that changes every batch --- any resulting failure is
trivially guaranteed rather than diagnostic.

We instead use \citet{theiler1992testing} surrogate data
testing in its standard form: train a causal SpectralQK to
convergence, then evaluate the \emph{frozen} model under
four perturbations applied only at evaluation time:
(i) \textsc{standard}, the unperturbed forward pass;
(ii) \textsc{phase\_rand}, a shared random phase applied to
both Q and K --- sharing the same random phase across Q and
K destroys absolute temporal alignment while preserving
relative Q--K phase coherence, the standard Theiler
surrogate construction;
(iii) \textsc{amp\_only}, replacing the filtered signal with
its amplitude envelope only ($\mathrm{irfft}(|\hat{Q}_f|\cdot w)$,
discarding all phase information while preserving the time
index); and
(iv) \textsc{phase\_shift}, a constant $\pi/4$ phase offset
applied to all frequencies (a control for sensitivity to
absolute, as opposed to relative, phase).
The DC and Nyquist bins are kept strictly real under all
perturbations to preserve the Hermitian symmetry required
for a real-valued inverse transform.

\begin{table}[h]
\centering
\caption{
  Theiler-style surrogate analysis, causal SpectralQK (bin=4,
  period=64 tokens), evaluated at the best training step.
}
\label{tab:theiler}
\begin{tabular}{lrl}
\toprule
Condition & Val & Interpretation \\
\midrule
BASE-DOT        & 1.474 & reference \\
Standard        & 1.498 & trained model \\
Phase shift ($\pi/4$) & 1.617 & control: absolute phase \\
Phase random (shared)  & 3.159 & Theiler surrogate \\
Amplitude only  & 4.097 & phase fully discarded \\
\bottomrule
\end{tabular}
\end{table}

Three findings follow. First, amplitude alone
(val\,=\,4.097) is far worse than even BASE-DOT
(1.474), confirming that phase coherence is not merely
additive but \emph{necessary}: the correct frequency band
without phase information actively harms attention.
Second, decomposing the surrogate range
($\Delta_{\text{amp}} = 4.097-1.498 = 2.599$) into
the phase-randomisation cost
($\Delta_{\text{phase}} = 3.159-1.498 = 1.661$,
$64\%$ of the range) and the residual spectral-selectivity
contribution ($0.938$, $36\%$) gives a quantitative
split: phase coherence accounts for roughly twice as
much of the spectral attention gain as the choice of
frequency band itself.
Third, the phase-shift control (val\,=\,1.617) is close
to standard (val\,=\,1.498), confirming the model is
largely invariant to a constant phase offset --- it
exploits the \emph{relative} phase between Q and K at
matching positions, not the absolute phase of either
signal alone, consistent with the cross-correlation
interpretation of attention developed in
\S\ref{sec:sa}.

\begin{figure}[t]
\centering
\includegraphics[width=\linewidth]{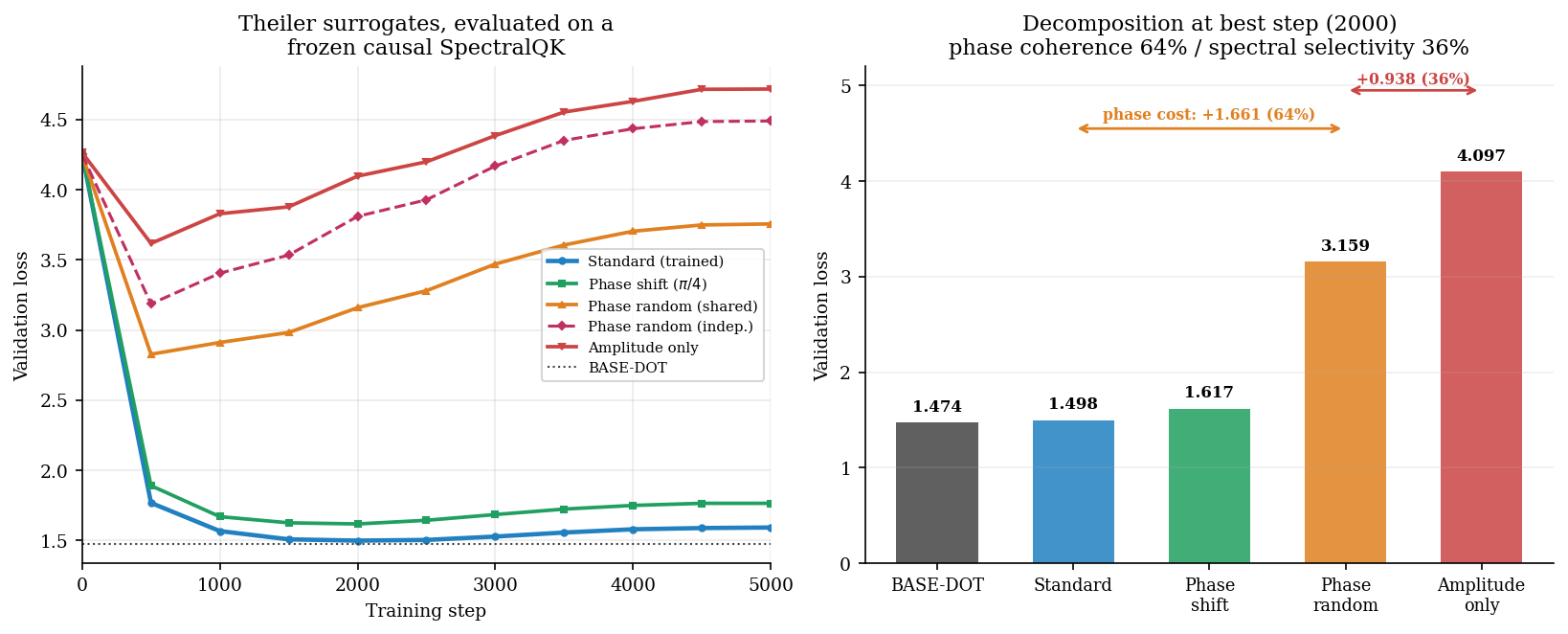}
\caption{
  Theiler-style surrogate analysis of a frozen, trained causal SpectralQK.
  \textbf{Left}: validation loss under each surrogate condition
  across training. Standard (blue) and phase-shift (green)
  track closely throughout, while phase-random (orange,
  pink) and amplitude-only (red) diverge from training step
  500 onward and continue to worsen as the model learns more
  from phase coherence.
  \textbf{Right}: the decomposition at the best training step
  (2000), with the phase cost (orange, 64\% of the surrogate
  range) and the residual spectral-selectivity contribution
  (red, 36\%) annotated.
}
\label{fig:theiler}
\end{figure}

This experiment is run on the causal (non-leaking)
implementation; the corresponding bilateral-FFT result in
the main results above necessarily reflects some
contribution from non-causal leakage (\S\ref{sec:morlet_negative}),
so the 64/36 split should be read as characterising the
\emph{mechanism} of spectral attention rather than as a
precise decomposition of the bilateral-FFT val values
reported in Table~\ref{tab:main}.

\paragraph{Phase synchrony as the attention criterion.}
Under this view, Fourier-QK detects recurring temporal
patterns at paragraph scale in the Q/K representation
space through a phase-synchrony principle:
two tokens receive a high attention score when they
occur at similar phase positions within the
paragraph-scale temporal rhythm of the text,
corresponding to structurally similar points
in the discourse.

\paragraph{Learned vs.\ fixed frequency.}
Fourier-QK ($+0.600$) outperforms Random-QK ($+0.443$).
The learned dominant frequency per head extracts
a task-specific spectral feature from the Q/K space.
Different heads may learn different dominant frequencies,
allowing multi-resolution spectral attention.

\paragraph{The bilateral requirement.}
All causal time-domain filters fail to improve over
BASE-DOT, while bilateral FFT filters succeed.
This suggests the spectral preprocessing benefit
requires access to the full sequence context in the
filter application --- a truly non-local operation.
This is consistent with the known effectiveness of
FNet's global Fourier mixing~\citep{lee2021fnet}.

\paragraph{Retrospective view on Papers 1--4.}
Paper~2~\citep{authorname2025phase4} found that
\EGA{} and \MoPE{} are superadditive ($+0.119$),
suggesting energy and phase are complementary signals.
The current results suggest the complementarity
comes from the spectral nature of both operations,
not from the specific Morlet wavelet interpretation.
Paper~3~\citep{authorname2025mope} observed boundary
saturation in learned \MoPE{} parameters; the current
finding that spectral preprocessing helps in the
bilateral setting may explain this: the optimal
positional encoding operates in the frequency domain.
Paper~6 of this series extends the spectral framework
to a state-space formulation, replacing the finite-context
FIR wavelet filter with an infinite-context IIR state
\citep{gu2021efficiently}, with the diagonal state
transition initialised from the near-dyadic frequency
hierarchy identified here.

\paragraph{Limitations.}
All experiments are at small scale
($\leq$6M parameters, character-level, $T=256$,
single seed for most comparisons).
The non-causal symmetric kernel problem prevents
any irfft-based position reconstruction from being
used in causal language models.
This includes Morlet, Mexican Hat, and Gaussian
filters --- all produce symmetric kernels.

\paragraph{Positional encoding and gain decomposition.}
All models use learned additive positional encoding:
$x = \text{tok}(t) + \text{pos}(t)$.
The Q/K projections therefore contain both semantic
content (token embeddings) and positional structure
(positional embeddings), and the Fourier-QK frequency
selector may exploit either or both components.

We tested Fourier-QK-Init4 with three PE variants:
learned (standard), no PE (zero), and fixed sinusoidal.
The Fourier-QK gain above the corresponding BASE-DOT
(same PE) decomposes cleanly:

\begin{center}
\begin{tabular}{lrrl}
\toprule
PE type & Init4 val & Gain vs BASE & Freq (bin) \\
\midrule
Learned  & 0.608 & $+0.865$ (100\%) & 4.7 \\
No PE    & 0.981 & $+0.527$ (61\%)  & 4.3 \\
Sinusoidal & 1.389 & $+0.137$ (16\%) & 3.2 \\
\bottomrule
\end{tabular}
\end{center}

\textbf{Component A (token content, 61\%):}
Fourier-QK retains $\Delta$\,=\,$+0.527$ with no positional
encoding, confirming that paragraph-scale spectral
structure ($\sim$60 tokens per cycle) exists in the
character sequence content itself, independent of
positional signals.
This is the core contribution.

\textbf{Component B (positional encoding, 39\%):}
The additional gain with learned PE suggests the
positional embedding develops spectral content
complementary to the paragraph-scale frequency
selector --- effectively acting as a spectral
amplifier at bin\,$\approx$\,4--5.

\textbf{Sinusoidal PE interferes with Fourier-QK:}
Surprisingly, sinusoidal PE (Vaswani et al.\ 2017)
performs much worse than no PE (gain $+0.137$ vs $+0.527$).
Sinusoidal PE has frequency components at very low
normalised frequencies ($\omega \sim 1/10000^{2i/d}$),
which are not at the paragraph scale.
The Fourier-QK selector drifts to bin\,=\,3.2 (period 80 tokens)
trying to align with the sinusoidal PE frequencies
rather than the genuine paragraph-scale optimum ---
an instance of spectral interference between the PE
and the attention mechanism.
This result motivates the use of RoPE~\citep{su2024rope}
in future work, as rotary encoding contributes no
additive frequency components to the Q/K FFT.
The only verified clean approach (Fourier-QK,
Random-QK) uses frequency-domain feature collapse
rather than position reconstruction.
The shuffled gap diagnostic confirms no leakage for
Fourier-QK and Random-QK but cannot rule out
all forms of spectral artifact.

\paragraph{Future work.}
The scale sweep (Table~\ref{tab:scales}) reveals a
monotonic improvement with scale: val improves from
1.754 at $a=2$ tokens to 1.586 at $a=32$ tokens,
with no sign of saturation.
This points directly to \textbf{word-level tokenisation}
as the natural next experiment.
At BPE tokenisation (avg $\sim$4 characters/token),
$a=8$ characters $\approx$ 2 word-tokens --- a scale
where morphological and lexical structure is richest.
The Nyquist constraint and aliasing problem that
prevent causal Morlet from working at character scale
largely disappear at word scale, where linguistic
units span 2--10 tokens.

Five directions constitute a natural follow-on paper:

\textbf{(1) Learnable frequency selector width.}
The current Gaussian selector uses a fixed width
$\sigma^2 = 4$ bins, chosen by convenience.
A learnable width per head
$w(\omega, f^*, \sigma) = \exp(-(\omega-f^*)^2/2\sigma^2)$
would allow the model to discover whether sharp
(narrow $\sigma$) or broad (wide $\sigma$) frequency
selection is optimal.
If learned $\sigma \to 0$, the result converges to
a hard single-bin selector (consistent with the
fixed bin=4 result matching the learned Init4).
If $\sigma$ remains large, the Gaussian is capturing
a genuine spectral band rather than a single frequency.
This also connects to the uncertainty principle:
the optimal width trades frequency precision against
temporal precision in the Q/K representation space.

\textbf{(2) Wavelet-collapse attention and admissibility.}
We tested four frequency-collapse filters at paragraph
scale --- all avoiding irfft reconstruction, hence no
leakage --- and found a striking ranking:

\begin{center}
\small
\begin{tabular}{lrrl}
\toprule
Filter & Val & $\Delta$ & Admissible? \\
\midrule
Morlet-Collapse  & 1.152 & $+0.322$ & No (small DC) \\
Gauss-Collapse   & 0.608 & $+0.867$ & No ($\hat\psi(0)\neq0$) \\
Paul-Collapse    & 0.385 & $+1.090$ & Yes (one-sided) \\
MexHat-Collapse  & 0.370 & $+1.105$ & Yes ($\hat\psi(0)=0$) \\
MultiFourier (4) & 0.286 & $+1.166$ & --- \\
\bottomrule
\end{tabular}
\end{center}

The filter shape ranking --- MexHat $>$ Paul $\gg$ Gauss $\gg$ Morlet
--- correlates with the sharpness of the spectral bandpass.
\section{Related Work}
\label{sec:related}

\paragraph{FNet.}
\citet{lee2021fnet} replaces attention with Fourier
mixing of embeddings, achieving 92--97\% of BERT.
Our work is distinct: we apply spectral preprocessing
to Q/K projections while preserving the attention
score structure.

\paragraph{Wavelet attention.}
\citet{verma2024waveletgpt} inject Haar wavelet
structure into intermediate embeddings between decoder
blocks, achieving a 40--60\% pretraining speedup at
no parameter cost.
Our approach differs by applying spectral filters
directly to the Q/K similarity computation rather
than to the residual stream, and by learning the
filter frequency rather than fixing it to the dyadic
Haar schedule.

\paragraph{Spectral methods in transformers.}
\citet{verma2024signal} apply causal convolutional
filter banks between transformer layers.
\citet{tamkin2020language} use DCT decomposition
for multi-scale representations.
Our work applies spectral filtering inside the
attention scoring mechanism.

\paragraph{Prior series papers.}
Papers~1--4~\citep{authorname2025ega,authorname2025phase4,
authorname2025mope,authorname2025pod} established
energy gating, Morlet positional encoding,
and multiscale POD as complementary spectral
inductive biases.
The current paper investigates spectral preprocessing
of Q/K as a unified approach to spectral attention.

\section{Conclusion}
\label{sec:conclusion}

We have shown that FFT-based bilateral spectral
preprocessing of learned Q/K projections genuinely
improves transformer attention on character-level
language modelling.
A random spectral filter achieves $+0.443$ over
standard attention; a model with a single learned
dominant frequency achieves $+0.600$.
Both results are verified by a shuffled validation
diagnostic confirming genuine sequence learning (evidenced by large shuffled gap).

Causal time-domain filters (Gaussian, Mexican Hat,
causal Morlet) do not improve over standard attention,
establishing that the bilateral FFT operation is
essential rather than local Q/K smoothing.

We report the failure of a Morlet wavelet
cross-correlation formulation with full transparency:
the bilateral FFT implementation suffers from
circular boundary leakage, while the causal
implementation suffers from aliasing at sub-token
scales.
Resolving this remains the primary challenge for
future causal spectral attention work.

\bibliographystyle{plainnat}

\appendix

\section{Implementation Details}
\label{app:impl}

\paragraph{Fourier-QK forward pass.}
\begin{enumerate}
  \item Compute Q/K projections: $q = W_Q x$,
        $k = W_K x$
  \item FFT along position: $\hat{q} = \mathcal{F}(q)$,
        $\hat{k} = \mathcal{F}(k)$
  \item Soft frequency selection via Gaussian weights
        over bins centred at learned $f^*$ per head
  \item irfft to reconstruct filtered
        $\tilde{q}(b)$, $\tilde{k}(b)$
  \item Score: $e_{ij} = \mathrm{Re}[
        \tilde{q}(i)^* \cdot \tilde{k}(j)] / \sqrt{d}$
  \item Causal mask, softmax, value aggregation
        (standard)
\end{enumerate}

\paragraph{Leakage diagnostic.}
At each evaluation step, compute loss on both the
standard validation set (ordered) and a once-shuffled
version (same tokens, random order).
The shuffled gap $= \ell_\text{shuffled} - \ell_\text{ordered}$
should be large ($>1.0$) for genuine sequence models
and near zero for leaking models.
BASE-DOT achieves gap $= +5.78$ at step 5000;
Fourier-QK achieves $+4.32$; Random-QK achieves $+4.98$.

\paragraph{Causal convolution.}
All causal filters use left-only padding:
\texttt{F.pad(x, (K-1, 0))} followed by
\texttt{F.conv1d} with kernel of $K$ taps.
This guarantees position $i$ sees only
positions $\leq i$.

\paragraph{Why irfft position reconstruction leaks.}
The filtered signal reconstructed via irfft at position $i$ is:
\begin{equation}
  W_q(i) = \sum_{t=0}^{T-1} q(t) \cdot \kappa(i-t \bmod T)
\end{equation}
where $\kappa(\tau) = \mathcal{F}^{-1}[\hat\phi](\tau)$
is the filter impulse response.
For any real-valued symmetric filter
$\hat\phi(\omega) \in \Re$, we have
$\kappa(-\tau) = \kappa(\tau)$,
so $W_q(i)$ depends on both past and future tokens.

Zero-padding to $2T$ eliminates the modular wrap
(position 0 seeing position $T-1$) but does not
change the bilateral support of $\kappa$:
\begin{equation}
  W_q(i) = \sum_{t=0}^{T-1} q(t) \cdot \kappa(i-t)
  + \underbrace{\sum_{t=T}^{2T-1} 0 \cdot \kappa(i-t)}_{=0}
\end{equation}
The zero-padded region contributes nothing, but the
bilateral kernel still couples position $i$ to all
positions $t < i$ and $t > i$.
Confirmed experimentally: Morlet zero-padded ($2T$)
gives val=0.023 at step 500, same as $T$-point FFT.

\paragraph{Why Fourier-QK does not leak.}
Fourier-QK computes:
\begin{equation}
  \tilde{q}(i) = \sum_\omega \hat{q}(\omega)
    \cdot w(\omega, f^*) \cdot e^{i2\pi\omega i/T}
\end{equation}
This is still a bilateral operation, but the Gaussian
weight $w(\omega, f^*)$ is applied in the frequency
domain \emph{before} any position-indexed
reconstruction.
The key difference: the weighted sum over $\omega$
produces a single complex number per position that
is dominated by the spectral content at $f^*$,
not a reconstruction that explicitly couples
neighbouring positions through a symmetric kernel.
The causal mask on the attention score matrix
is then sufficient to prevent future information
from affecting past positions in the final output.

\end{document}